\documentclass{article}
\usepackage{ijcai20}

\usepackage{times}
\usepackage{soul}
\usepackage{url}
\usepackage[hidelinks]{hyperref}
\usepackage[utf8]{inputenc}
\usepackage[small]{caption}
\usepackage{graphicx}
\usepackage{amsmath}
\usepackage{amsthm}
\usepackage{booktabs}
\usepackage{algorithm}
\usepackage{algorithmic}
\usepackage{multirow}
\usepackage{amsfonts}
\usepackage{fancyhdr}
\renewcommand\footnotemark{}

\title{Evolving Robust Neural Architectures to Defend from Adversarial Attacks} 

\author{
Shashank Kotyan \And Danilo Vasconcellos Vargas
\affiliations
Department of Informatics, Kyushu University, Japan
\emails
shashankkotyan@gmail.com \And vargas@inf.kyushu-u.ac.jp
}

\begin{document}

\maketitle

\begin{abstract}
Neural networks are prone to misclassify slightly modified input images.
Recently, many defences have been proposed, but none have improved the robustness of neural networks consistently.
Here, we propose to use adversarial attacks as a function evaluation to search for neural architectures that can resist such attacks automatically.
Experiments on neural architecture search algorithms from the literature show that although accurate, they are not able to find robust architectures.
A significant reason for this lies in their limited search space.
By creating a novel neural architecture search with options for dense layers to connect with convolution layers and vice-versa as well as the addition of concatenation layers in the search, we were able to evolve an architecture that is inherently accurate on adversarial samples. 
Interestingly, this inherent robustness of the evolved architecture rivals state-of-the-art defences such as adversarial training while being trained only on the non-adversarial samples.
Moreover, the evolved architecture makes use of some peculiar traits which might be useful for developing even more robust ones.
Thus, the results here confirm that more robust architectures exist as well as opens up a new realm of feasibilities for the development and exploration of neural networks.
\end{abstract}

\section{Introduction}

    Neural Architecture Search (NAS) and adversarial samples have rarely appeared together.
    Regarding adversarial samples, they were discovered in $2013$ when neural networks were shown to behave strangely for nearly the same images \cite{szegedy2014intriguing}.
    Afterwards, a series of vulnerabilities were found in \cite{moosavi2017universal,su2019one}. 
    Such adversarial attacks can also be easily applied to real-world scenarios which confer a big problem for current deep neural networks' applications. 
    Currently, there is not any known learning algorithm or procedure that can defend against adversarial attacks \textit{consistently}.

    Regarding NAS, the automatic design of architectures has been of broad interest for many years.
    The aim is to develop methods that do not need specialists in order to be applied to a different application.
    This would confer not only generality but also easy of use. 
    Most of the algorithms for NAS are either based on reinforcement learning \cite{pham2018efficient,zoph2018learning} 
    or evolutionary computation \cite{real2017large,miikkulainen2019evolving}. 
    On the one hand, in reinforcement learning approaches, architectures are created from a sequence of actions which are afterwards rewarded proportionally to the crafted architecture's accuracy. 
    On the other hand, in evolutionary computation based methods, small changes in the architecture (mutations) and recombinations (crossover) are used to create new architectures. 
    All architectures evolved are evaluated based on their accuracy. 
    Some of the best architectures based on this accuracy are chosen to continue to the next generation.

    Here we propose the use of NAS to tackle the robustness issues exposed by adversarial samples.
    In other words, architecture search will be employed not only to find accurate neural networks but also robust ones.
    This is based on the principle that robustness of neural networks can be assessed by using accuracy on adversarial samples as an evaluation function. 
    We hypothesise that if there is a solution in a given architecture search space, the search algorithm would be able to find it.
    This is not only a blind search for a cure.
    The best architectures found should also hint which structures and procedures provide robustness for neural networks.
    Therefore, it would be possible to use the results of the search to understand further how to improve the representation of models as well as design yet more robust ones. 

\section{Adversarial Machine Learning}

    Adversarial machine learning is a constrained optimisation problem.
    Let $f(x) \in [\![1..N]\!]$ be the output of a machine learning algorithm in multi-label classification setting. 
    Here, $x \in \mathbb{R}^{k}$ is the input of the algorithm for the input of size $k$ and $N$ is the number of classes in which $x$ can be classified.
    In Image Classification problem $k = m \times n \times 3$ where $m \times n$ is the the size of the image.
    Adversarial samples $x'$ can be thus defined as follows:
    \begin{equation}  \begin{aligned}
    x' = x + \epsilon_{x} \quad \text{such that} \quad f(x') \ne f(x)
    \end{aligned} \end{equation} 
    in which $\epsilon_{x} \in \mathbb{R}^{k}$ is a small perturbation added to the input.
    Therefore, adversarial machine learning can be defined as an optimization problem\footnote{Here the definition will only concern untargeted attacks but a similar optimization problem can be defined for targeted attacks}:
    \begin{equation} \begin{aligned}
    \underset{\epsilon_{x}}{\text{minimize}} \quad g(x+\epsilon_{x})_c \quad \text{subject to} \quad \Vert \epsilon_{x} \Vert \leq th
    \label{adv_eqn}
    \end{aligned} \end{equation}
    where $th$ is a pre-defined threshold value and $g()_c$ is the soft-label or confidence for the correct class $c$ such that $f(x) = \text{argmax } g(x)$

    Moreover, attacks can be divided according to the function optimised.
    In this way, there are $L_0$ (limited number of pixels attacked), $L_1$, $L_2$ and $L_\infty$ (limited amount of variation in each pixel) types of attacks.
    There are many types of attacks as well as their improvements.
    Universal perturbation types of attacks were shown possible in which a single perturbation added to most of the samples is capable of fooling a neural network in most of the cases \cite{moosavi2017universal}. 
    Moreover, extreme attacks such as only modifying one pixel ($L_0 = 1$) called one-pixel attack is also shown to be surprisingly effective \cite{su2019one,vargas2019robustness}.
    Most of these attacks canbe easily transferred to real scenarios by using printed out versions of them \cite{kurakin2016adversarial}. 
    Moreover, carefully crafted glasses \cite{sharif2016accessorize} or even general 3D adversarial objects are also capable of causing misclassification \cite{athalye2017synthesizing}. 
    Regarding understanding the phenomenon, it is argued in \cite{goodfellow2014explaining} that neural networks' linearity is one of the main reasons.
    Another recent investigation proposes the conflicting saliency added by adversarial samples as the reason for misclassification \cite{vargas2019understanding}.

    Many defensive systems were proposed to mitigate some of the problems.
    However, current solutions are still far from solving the problems.
    Defensive distillation \cite{papernot2016distillation} uses a smaller neural network to learn the content from the original one; however, it was shown not to be robust enough \cite{carlini2017towards}.
    The addition of adversarial samples to the training dataset, called adversarial training, was also proposed \cite{goodfellow2014explaining,huang2015learning,madry2017towards}. 
    However, adversarial training has a strong bias in the type of adversarial samples used and is still vulnerable to attacks 
    Many recent variations of defences were proposed which are carefully analysed, and many of their shortcomings explained in \cite{athalye2018obfuscated,uesato2018adversarial}. 
    In this article, different from previous approaches, we aim to tackle the robustness problems of neural networks by automatically searching for inherent robust architectures.

\section{Neural Architecture Search}

    \begin{figure*}[!htb]
        \centering
        \includegraphics[width=0.75\linewidth]{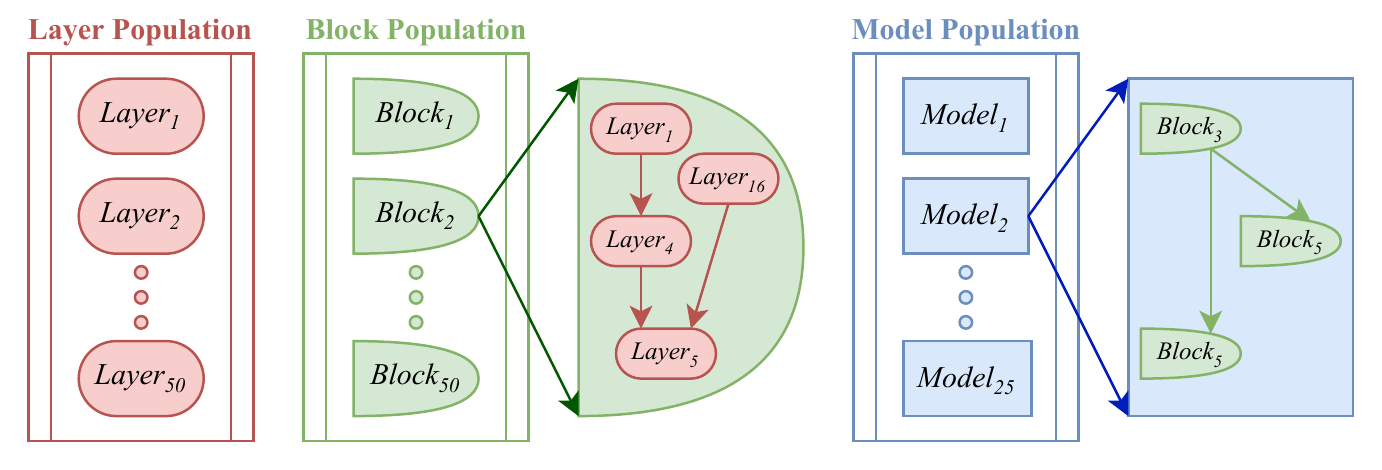}
        \caption{Illustration of the proposed RAS structure with three subpopulations.}
        \label{arch} 
    \end{figure*}

    There are three components to a neural architecture search: search space, search strategy and performance estimation. 
    A search space substantially limits the representation of the architecture in a given space. 
    A search strategy must be employed to search for architectures in a defined search space.
    Some widely used search strategies for NAS are:  
    Random Search, 
    Bayesian Optimization, 
    Evolutionary Methods, 
    Reinforcement Learning, 
    and Gradient Based Methods. 
    Finally, a performance estimation (usually error rate) is required to evaluate the explored architectures. 

    Currently, most of the current NAS suffer from high computational cost while searching in a relatively small search space \cite{lee2018hardness,lima2019evolving}.
    It is already shown in \cite{yu2018worth} that, if there is a possibility of fitness approximation at small search spaces, we could evolve algorithms in an ample search space. 
    Moreover, many architecture searches focus primarily on the hyper-parameter search while using architecture search spaces around previously hand-crafted architecture \cite{lee2018hardness,he2019automl} such as DenseNet which are proved to be vulnerable to adversarial attacks \cite{vargas2019robustness}. 
    Therefore, for finding robust architectures, it is crucial to expand the search space beyond the current NAS.

    SMASH \cite{brock2017smash} uses a neural network to generate the weights of the primary model. 
    The main strength of this approach lies in preventing high computational cost, which is incurred in other searches. 
    However, this comes at the cost of not being able of tweaking hyperparameters which affect weights like initialisers and regularisers.
    Deep Architect \cite{negrinho2017deeparchitect} follows a hierarchical approach using various search algorithms such as Monte Carlo Tree Search (MCTS) and Sequential Model based Global Optimization (SMBO). 

\section{Searching for Robust Architectures}

    A robust evaluation (defined in Section \ref{rob_eval}) and search algorithm must be defined to search for robust architectures.
    The search algorithm may be a NAS provided that some modifications are made (Section \ref{rob_conv}).
    However, to allow for a more extensive search space, which is better suited to the problem, we also propose the Robust Architecture Search (Section \ref{ras_sec}).

    \subsection{Robustness Evaluation}
    \label{rob_eval}

        \begin{table*}[!htb]
        \centering
        \resizebox{0.8\linewidth}{!}{%
        \begin{tabular}{ll|cccccccc|r}
        \toprule
        \multirow{2}{*}{\textbf{Model}}    & \multirow{2}{*}{\textbf{Attack Optimiser}} & \multicolumn{4}{c}{\textbf{$L_0$ Attack}} & \multicolumn{4}{c|}{\textbf{$L_\infty$ Attack}} & \multirow{2}{*}{\textbf{\textit{Total}}}  \\
                 &           & $th=1$  & $th=3$  & $th=5$  & $th=10$                & $th=1$  & $th=3$  & $th=5$  & $th=10$ & \\
        \midrule
        \multirow{2}{*}{CapsNet} & DE     & 18 & 46 & 45 & 47 & 05 & 09 & 12 & 24 & \textit{206} \\
                                 & CMA-ES & 14 & 34 & 45 & 62 & 09 & 38 & 74 & 98 & \textit{374} \\
        \multirow{2}{*}{ResNet}  & DE     & 23 & 66 & 75 & 77 & 06 & 22 & 46 & 78 & \textit{393} \\
                                 & CMA-ES & 11 & 49 & 63 & 77 & 28 & 72 & 75 & 83 & \textit{458} \\
        \multirow{2}{*}{AT}      & DE     & 23 & 59 & 63 & 66 & 00 & 02 & 03 & 06 & \textit{222} \\
                                 & CMA-ES & 20 & 50 & 70 & 82 & 03 & 12 & 25 & 57 & \textit{319} \\
        \multirow{2}{*}{FS}      & DE     & 21 & 73 & 78 & 78 & 04 & 21 & 45 & 78 & \textit{398} \\
                                 & CMA-ES & 17 & 49 & 69 & 78 & 26 & 63 & 66 & 74 &\textit{ 442} \\
        \midrule
        \multicolumn{2}{r|}{\textbf{\textit{Total}}} & \textit{147} & \textit{426} & \textit{508} & \textit{567} & \textit{81} & \textit{239} & \textit{346}& \textit{ 498} & \textit{2812} \\
        \bottomrule
        \end{tabular}
        }
        \caption{
            The number of samples used from each type of black-box attack to compose the \textit{$2812$} adversarial samples.
            Based on the principle of the transferability of adversarial samples, these adversarial samples are used as a fast attack for the robustness evaluation of architectures.
            Details of the attacks as well as the motivation for using a model-agnostic (black-box) dual quality ($L_0$ and $L_\infty$) assessment are explained in detail at \protect\cite{vargas2019robustness}.
        }
        \label{adv_samples}
        \end{table*}
        
        Adversarial accuracy may seem like a natural evaluation function for assessing neural networks' robustness.
        However, there are many types of perturbations possible; each will result in a different type of robustness assessment and evolution.
        For example, let us suppose an evaluation with $th=5$ is chosen, robust networks against $th=5$ might be developed. At the same time, nothing can be said for other $th$ and attack types (different $L$).
        Therefore, $th$ plays a role but the different types of $L_0$, $L_1$, $L_2$ and $L_\infty$ completely change the type of robustness, such as wide perturbations ($L_\infty$), punctual perturbations ($L_0$) and a mix of both ($L_1$ and $L_2$).
        To avoid creating neural networks that are only robust against one type of robustness and at the same time to allow robustness to slowly build-up from any partial robustness, a set of adversarial attacks for varying $th$ and $L$ are necessary.

        To evaluate the robustness of architectures in varying $th$ and $L$ while at the same time keeping computational cost low, we use here a transferable type of attack.
        In other words, adversarial samples previously found by attacking other methods are stored and used as possible adversarial samples to the current model under evaluation.
        This solves the problem that most of the attacks are usually slow to be put inside a loop which can make the search for architectures too expensive.
        
        Table \ref{adv_samples} shows a summary of the number of images used from each type of attack, totalling $2812$ adversarial samples. 
        Samples were generated using the model agnostic dual quality assessment \cite{vargas2019robustness}.
        Specifically, we use the adversarial samples from two types of attacks ($L_0$ and $L_\infty$ attacks) with two optimization algorithms (Covariance Matrix Adaptation Evolution Strategy (CMA-ES) \cite{hansen2003reducing} and Differential Evolution (DE) \cite{storn1997differential}). 
        We use CIFAR-10 dataset \cite{krizhevsky2009learning} to generate the adversarial samples.
        We attacked traditional architectures such as ResNet \cite{he2016deep} and CapsNet \cite{sabour2017dynamic}.
        We also attacked some state-of-art-defences such as Adversarial Training (AT) \cite{madry2017towards} and Feature Squeezing (FS) \cite{xu2017feature} defending ResNet.
        The evaluation procedure consists of calculating the amount of successful adversarial samples divided by the total of possible adversarial samples.
        This also avoids problems with different amount of perturbation necessary for attacks to succeed, which could cause incomparable results.

    \subsection{Robust Search Conversion of existing NAS}
    \label{rob_conv}

        By changing the fitness function (in the case of evolutionary computation based NAS) or the reward function (in the case of reinforcement learning based NAS), it is possible to create robust search versions of NAS algorithms.
        In other words, it is possible to convert the search for accuracy into the search for robustness and accuracy.
        Here we use SMASH and DeepArchitect for the tests.
        The reason for the choice lies in the difference between the methods and availability of the code.
        Both methods have their evaluation function modified to contain not only accuracy but also robustness (Section \ref{rob_eval}).

\section{Robust Architecture Search (RAS)}
\label{ras_sec}
    
    \begin{figure*}[!htb]
            \centering
            \includegraphics[width=0.8\linewidth]{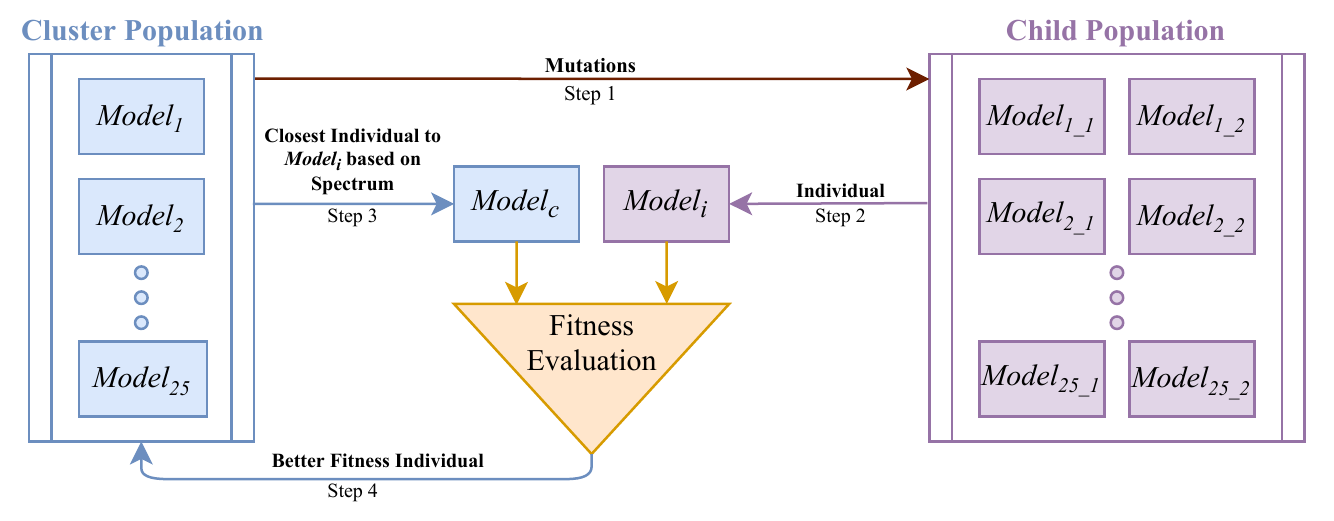} 
            \caption{
            Illustration of the proposed Evolutionary Strategy.
            In this strategy, Step 2, 3, and 4 are repeated for all the individuals in the child population. Step 1 is repeated when all the individuals in the child population have been evaluated against a cluster individual.}
            \label{es} 
    \end{figure*}

    Here, we propose an evolutionary algorithm to search for robust architectures called Robust Architecture Search (RAS)\footnote{Code is available at \url{http://bit.ly/RobustArchitectureSearch}}.
    It makes sense to focus here on search spaces that allow for unusual layer types and their combinations to happen, which is more vast than the current traditional search spaces.
    The motivation to consider this vast search space is that some of the most robust architectures might contain these unusual combinations which are not yet found or deeply explored.
   
    \paragraph{RAS Overview} RAS works by creating three initial populations (layer, block and model populations).
    Every generation, the model population have each of its members modified five times by mutations.
    The modified members are added to the population as new members.
    Here we propose a utility evaluation in which layer and block populations are evaluated by the number of models (architectures) using them.
    Models are evaluated by their accuracy and attack resilience (accuracy on adversarial samples).
    All blocks and layers which are not used by any of the current members of the model population are removed at the end step of each generation.
    Moreover, architectures compete with similar ones in their subpopulation, such that only the fittest of each subpopulation survives.

    \subsection{Description of Population}


        For such a vast search space to be more efficiently searched, we propose to use three subpopulations, allowing for the reuse of blocks and layers.
        Specifically, the layers consist of:
        \textit{Layer Population:} Raw layers (convolutional and fully connected) which make up the blocks. 
        \textit{Block Population:} Blocks which are a combination of layers.
        \textit{Model Population:} A population of architectures which consists of interconnected blocks.
        Figure \ref{arch} illustrates the architecture.
        
        The initial population consists of $25$ random architectures which contain $U(2,5)$ blocks made up of $U(2,5)$ layers, in which $U(a,b)$ is a uniform random distribution with minimum $a$ and maximum $b$ values.
        The possible available parameters for the layers are as follows: 
        for convolutional layers, filter size might be $8$, $16$, $32$ or $64$, stride size may be $1$ or $2$ and kernel size is either $1$, $3$ or $5$;
        for fully connected layers, the unit size may assume the values of $64$, $128$, $256$ or $512$.
        All the layers use Rectified Linear Unit (ReLU) as an activation function and are followed by a batch-normalisation layer. 

    \subsection{Mutation Operators}


        Regarding the mutation operators used to evolve the architecture, they can be divided into layer, block and model mutations which can only be applied to the respective layer, block and model populations' individuals. 
        The following paragraphs define the possible mutations.
        
        \paragraph{Layer Mutation} Layer mutations are of the following types:
        \textit{a) Change Kernel:} Changes the kernel size of the convolution layer, 
        \textit{b) Change Filter:} Changes the filter size of the convolution layer, 
        \textit{c) Change Units:}  Changes the unit size of the fully connected layer,
        \textit{d) Swap Layer:}    Chosen layer is swapped with a random layer from the layer population.
        
        \paragraph{Block Mutation} Block mutation change a single block in the block population. 
        The possibilities are: 
        \textit{e) Add Layer:}               A random layer is added to a chosen random block,
        \textit{f) Remove Layer:}            A random layer is removed from a chosen random block, 
        \textit{g) Add Layer Connection:}    A random connection between two layers from the chosen random block is added,
        \textit{h) Remove Layer Connection:} A random connection between the two layers from the chosen random block is removed,
        \textit{i) Swap Block:}              Chosen block is swapped with a random block from the population. 
        
        \paragraph{Model Mutation} Model mutation modify a given architecture.
        The possible model mutations are: 
        \textit{j) Add Block:}               A random block is added to the model,
        \textit{k) Remove Block:}            A random block is removed from the model,
        \textit{l) Add Block connection:}    A random connection between the two blocks is added,
        \textit{m) Remove Block connection:} A random connection between the two blocks is removed.

        All mutations add a new member to the population instead of substituting the previous one.
        In this manner, if nothing is done, the population of layers and blocks may explode, increasing the number of lesser quality layers and blocks.
        This would cause the probability of choosing functional layers and blocks to decrease.
        To avoid this, when the layer or block population exceeds $100$ individuals, the only layer/block mutation available is swap layers/blocks.

    \subsection{Objective (Fitness) Function}


        Fitness of an individual of the model population is measured using the final validation accuracy of the model after training for a maximum of $200$ epochs with early stopping if accuracy or validation accuracy do not change more than $0.001$ in the span of $15$ epochs. 
        Regarding the fitness calculation, the fitness is calculated as the accuracy of the model plus the robustness of the model $(\textbf{Fitness} = \textbf{Accuracy} + \textbf{Robustness})$.
        
        The $\textbf{Accuracy}$ of the architecture is calculated after the model is trained for $50$ epochs over the whole set of samples ($50000$ samples) of the CIFAR-10's training dataset for every $10$ generation(s) or over $1000$ random samples of the CIFAR-10 training dataset for all other generations. 
        This allows an efficient evolution to happen in which blocks and layers evolve at a faster rate without interfering with the architecture's accuracy.
        Using entire dataset subjects to evolving the architecture to have better accuracy and using a subset of the dataset evolves the layers and blocks of the architecture at a faster rate.
        The $\textbf{Robustness}$ of the architecture is calculated using accuracy on adversarial samples as described in Section \ref{rob_eval}.

    \subsection{Spectrum-based Niching Scheme}


        To keep a high amount of diversity while searching in a vast search space by using a novel algorithm described below also shown in Figure \ref{es}. 
        This niching scheme uses the idea of Spectrum-based niching from \cite{vargas2017spectrum} but explores a different approach to it.
        First, all the initial population is converted into a cluster population such that each individual in the initial population is a cluster representative. 
        Then we create two child individuals for each cluster representative by randomly applying five mutation operators on cluster representative. 
        We then find the closest cluster representative to the child individual using spectrum described below.
        If the fitness of the child individual is better than the closest cluster representative than the child individual becomes the new cluster representative, and the old cluster representative is removed from the population and the generation. 
        The process is completed for all the individuals in a cluster population. 
        We are hence evolving a generation of the evolution.


        Here, we use the spectrum as a histogram containing the features: Number of Blocks, Number of Total Layers, Number of Block Connections, Number of Total Layer Connections, Number of Dense Layers, Number of Convolution Layers, Number of Dense to Dense Connections, Number of Dense to Convolution Connections, Number of Convolution to Dense Connections, and Number of Convolution to Convolution Connections. 
        By using this Spectrum-based niching scheme, we aim to achieve an open-ended evolution, preventing the evolution from converging to a single robust architecture. 
        Preserving diversity in the population ensures that the exploration rate remains relatively high, allowing us to find different architectures even after many evolution steps.
        For the vast search space of architectures, this property is especially important, allowing the algorithm to traverse the vast search space efficiently.
    
\section{Experiments on RAS and Converted NAS}
    
    \begin{table}[!htb]
        \centering
        \resizebox{\linewidth}{!}{%
        
        \begin{tabular}{lcc}
            \toprule
            Architecture Search & Testing ER & ER on Adversarial Samples \\
            \midrule
            \protect DeepArchitect*  & 25\% & 75\% \\
            \protect SMASH*          & 23\% & 82\% \\
            \midrule
            Ours                     & \bf{18}\% & \bf{42}\% \\
            \bottomrule
        \end{tabular}
        }
        \caption{
            Error Rate (ER) on both the testing dataset and adversarial samples when the evaluation function has both accuracies on the testing data and accuracy on the adversarial samples. \\
            *Both DeepArchitect and SMASH had their evaluation function modified to be the sum of accuracy on the testing and adversarial samples. 
        }
        \label{smash}

    \end{table}

    Here, experiments are conducted on both the proposed RAS and converted versions of DeepArchitect and SMASH.
    The objective is to achieve the highest robustness possible using different types of architecture search algorithms and compare their result and effectiveness.
    Initially, DeepArchitect and Smash found architectures which had an error rate of $11\%$ and $4\%$ respectively when the fitness is only based on the neural network's testing accuracy.
    However, when the accuracy on adversarial samples is included in the evaluation function, the final error rate increases to $25\%$ and $23\%$ respectively (Table \ref{smash}).
    This may also indicate that poisoning the dataset might cause a substantial decrease in accuracy for the architectures found by SMASH and DeepArchitect.
    In the case of RAS, even with a more extensive search space, an error rate of $18\%$ is achieved.

    Regarding the robustness of the architectures found, Table \ref{smash} shows that the final architecture found by DeepArchitect and SMASH were very susceptible to attacks, with error rate on adversarial samples of $75\%$ and $82\%$ respectively.
    Despite the inclusion of the $R$ (measured accuracy on adversarial samples) on the evaluation function, the architectures were still unable to find a robust architecture.
    This might be a consequence of the relatively small search space used and more focused initialisation procedures.
    Moreover, the proposed method (RAS) finds an architecture which has an error rate of only $42\%$ on adversarial samples.
    \textit{ 
    Note, however, that in the case of the evolved architecture, this is an inherent property of the architecture found}.
    The architecture is inherently robust without any kind of specialised training or defence such as adversarial training (i.e., the architecture was only trained on the training dataset).
    The addition of defences should increase its robustness further.

\section{Analyzing RAS}
    
    \begin{figure*}[!htb]
        \centering
        \includegraphics[width=0.5\linewidth]{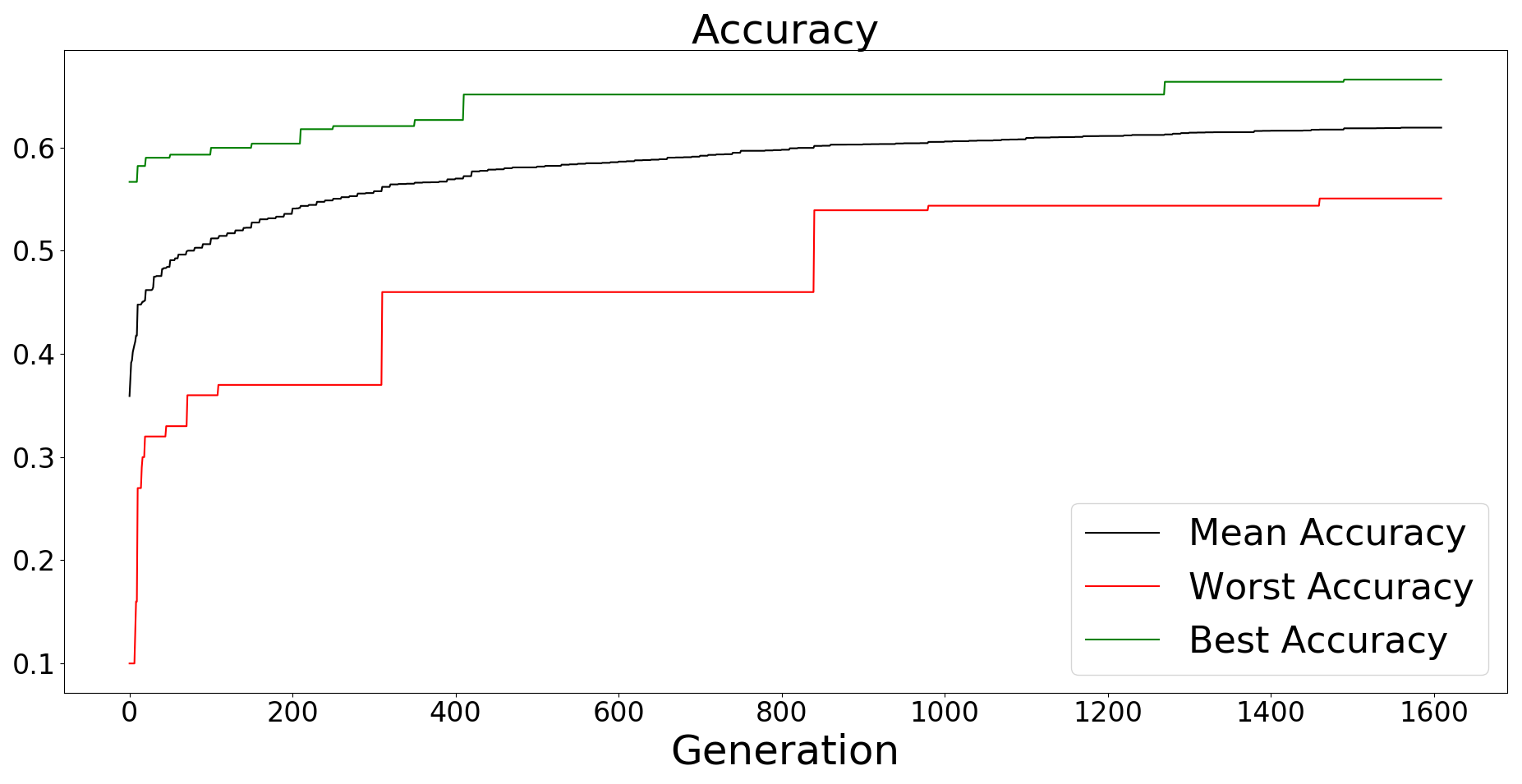} 
        \caption{Accuracy improvement over the generations.}
        \label{evolvable} 
    \end{figure*}

    \begin{figure*}[!htb]
        \centering
        \includegraphics[width=0.3\linewidth]{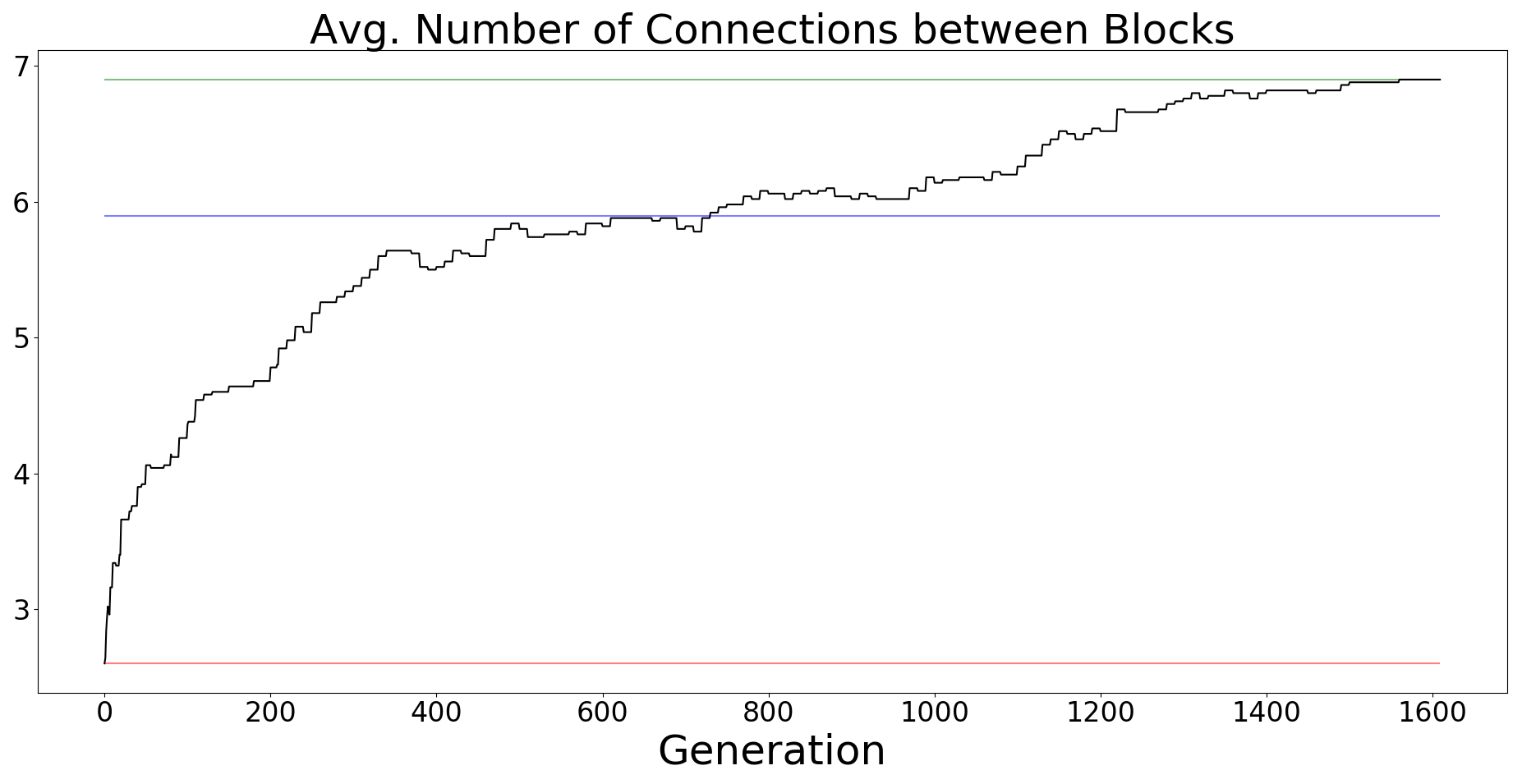} 
        \includegraphics[width=0.3\linewidth]{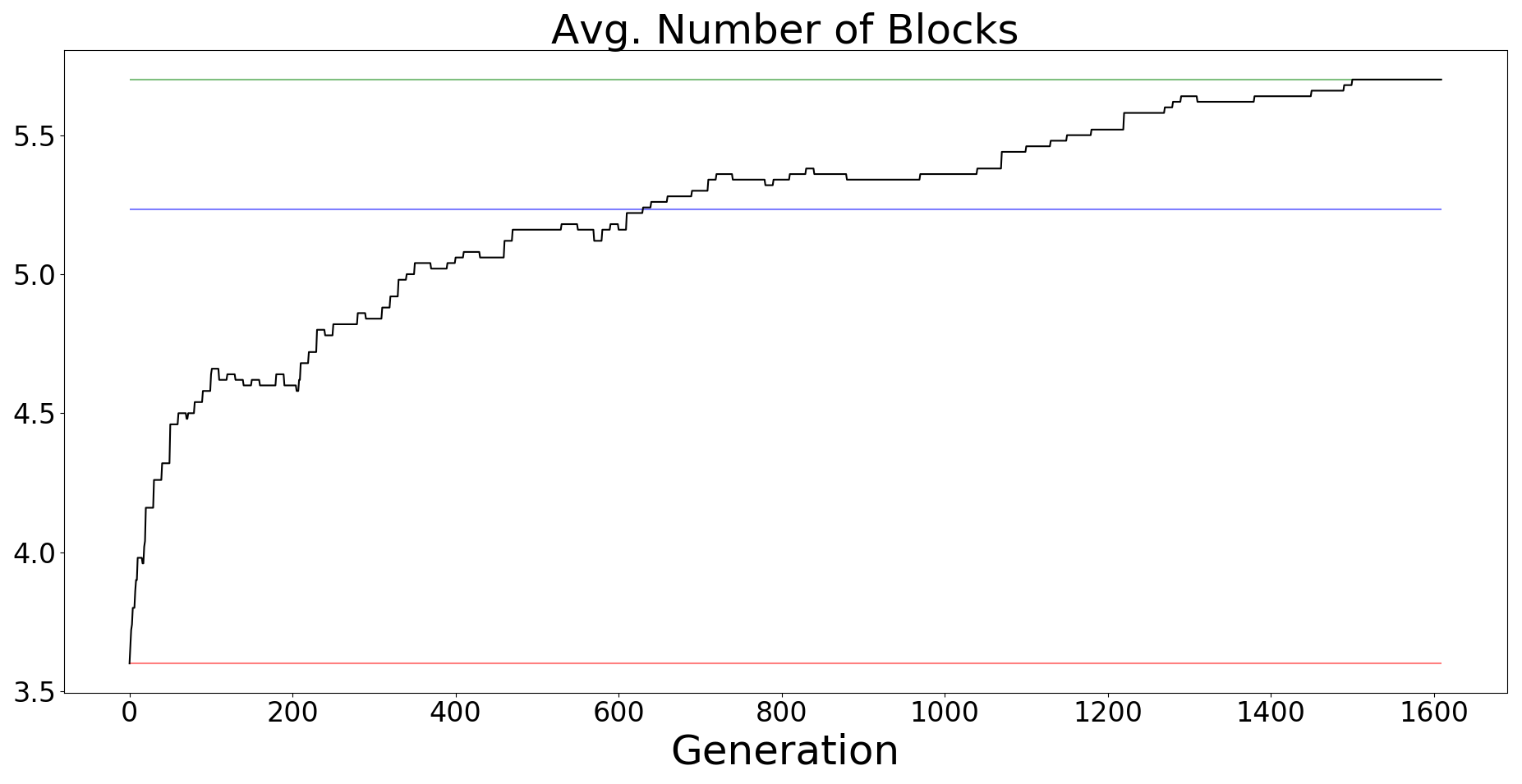} 
        \includegraphics[width=0.3\linewidth]{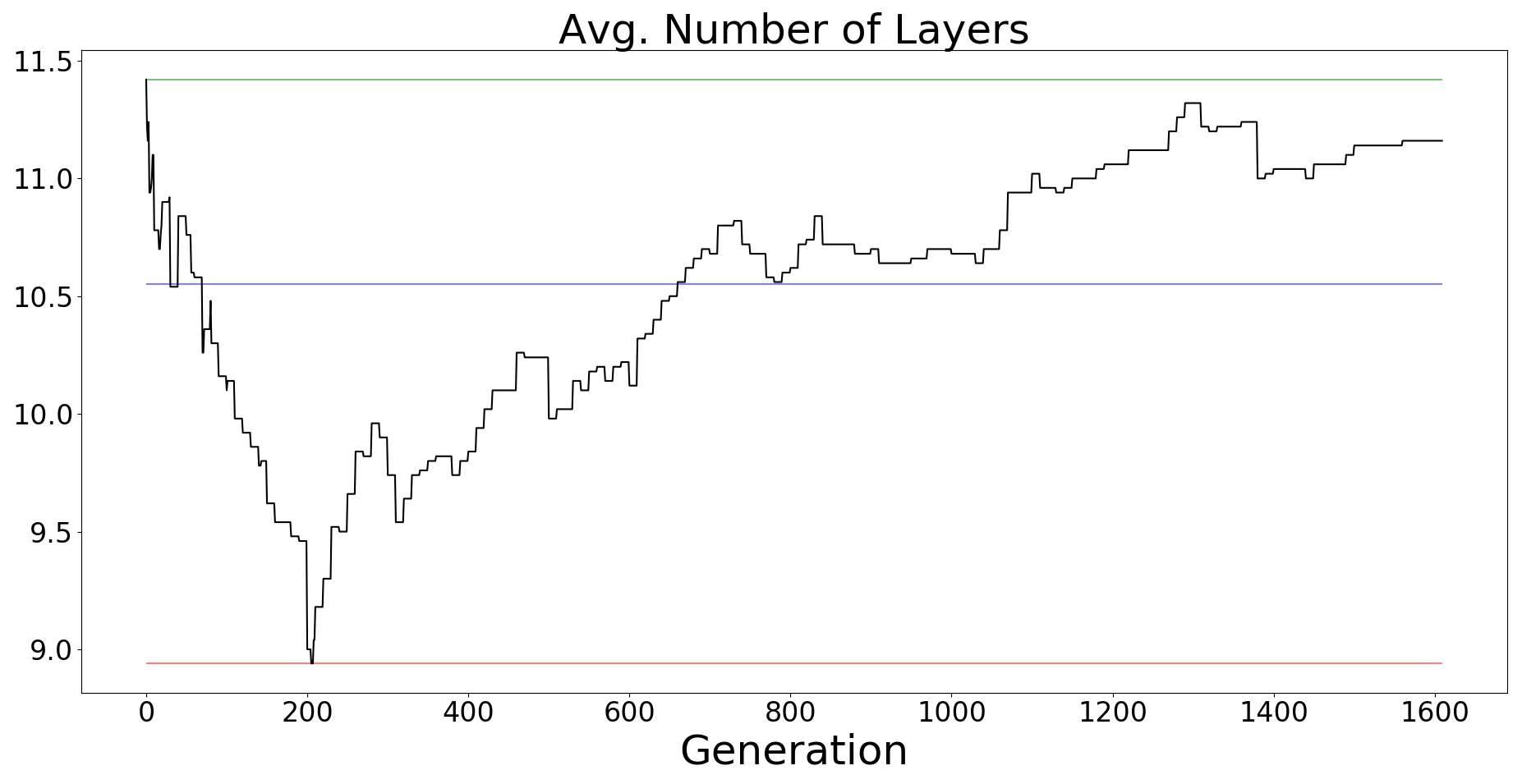} 
        \includegraphics[width=0.3\linewidth]{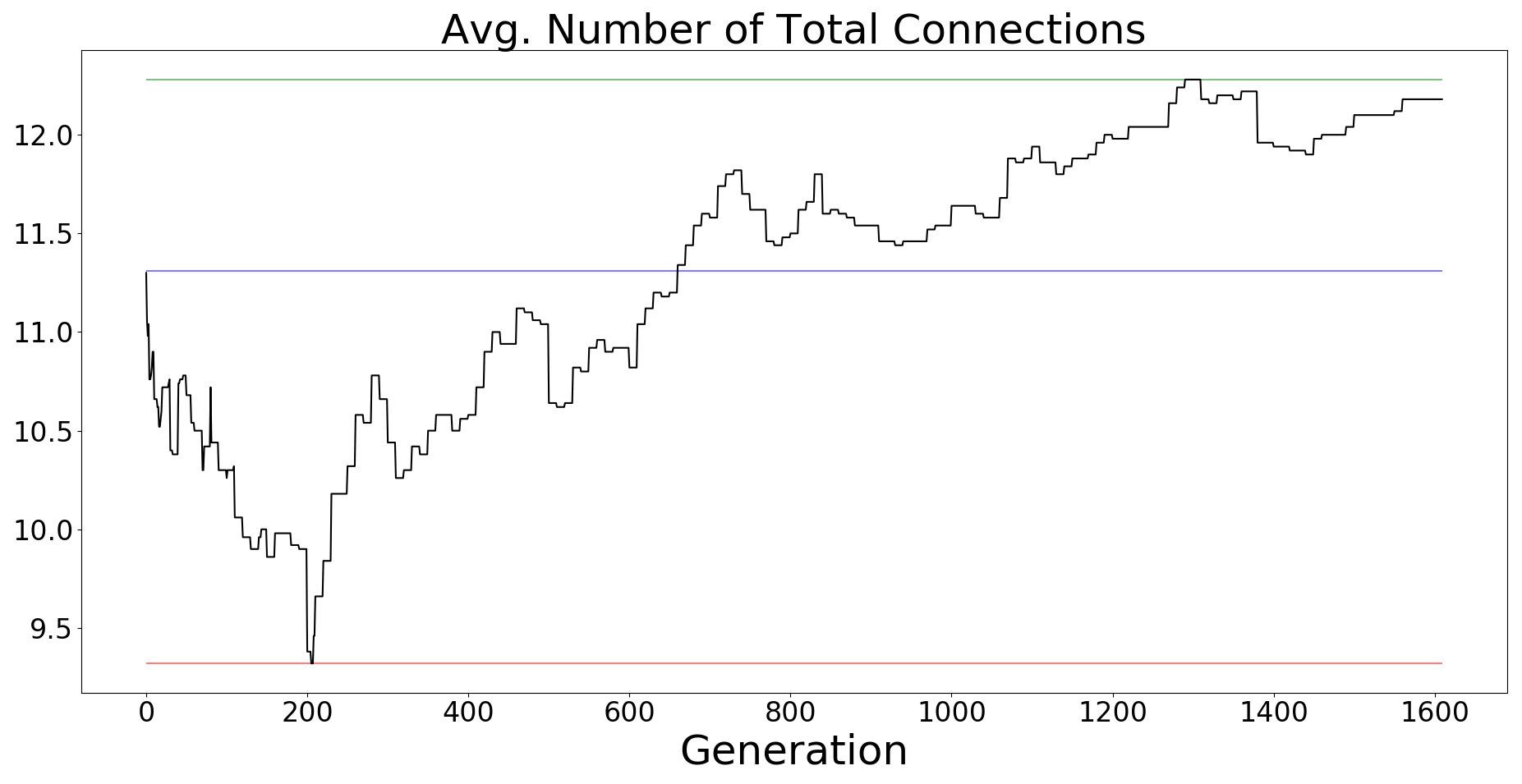} 
        \includegraphics[width=0.3\linewidth]{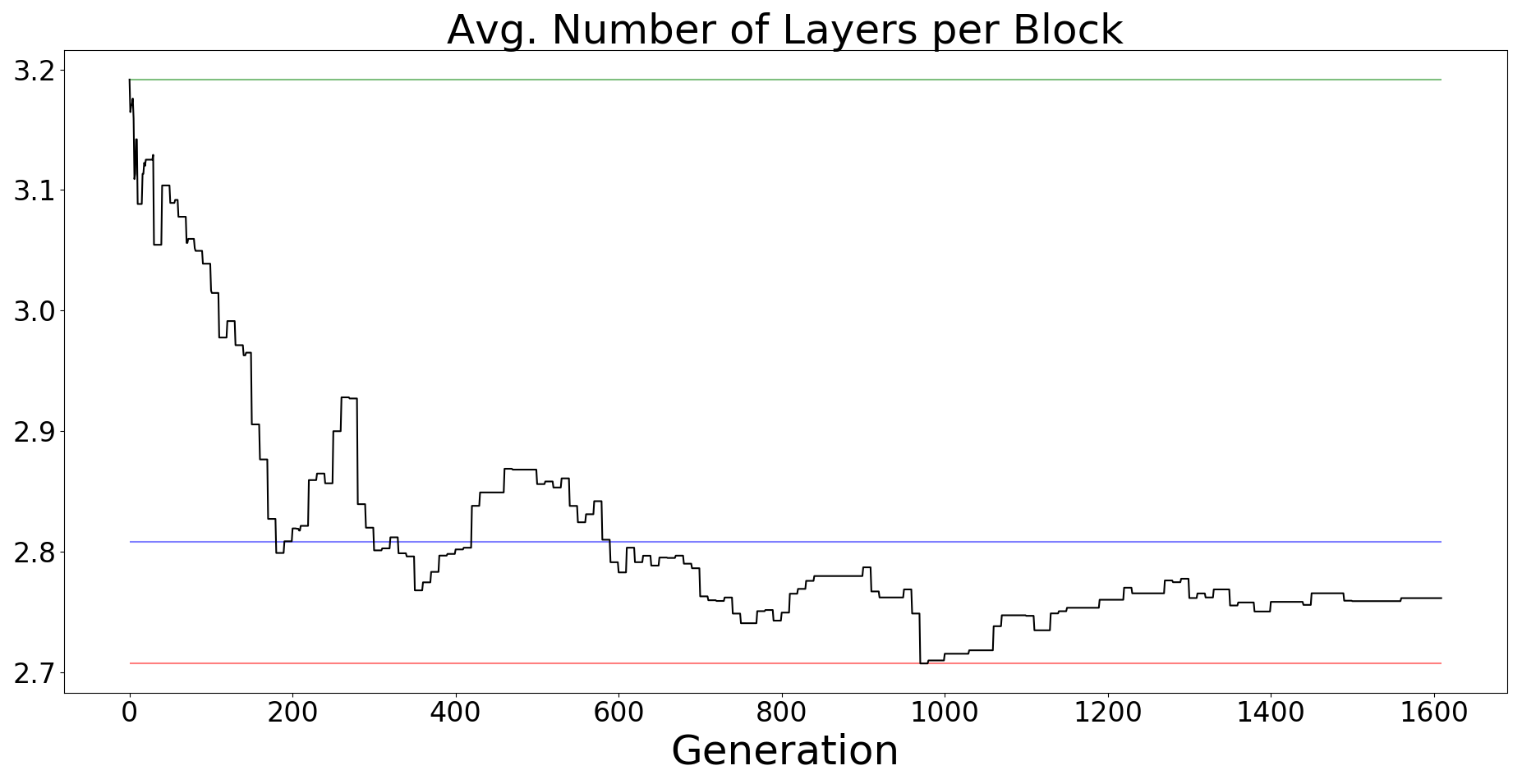} 
        \includegraphics[width=0.3\linewidth]{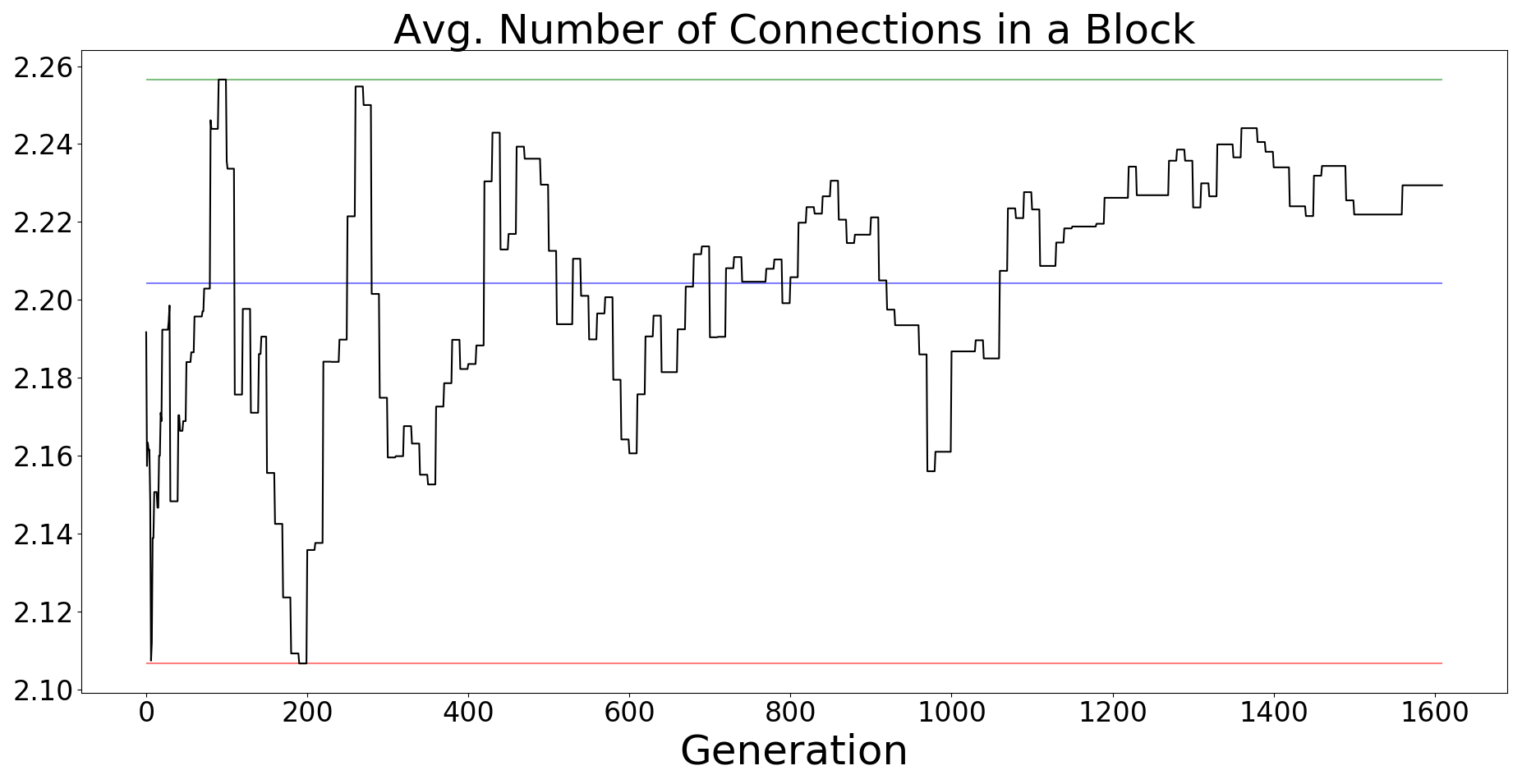} 
        \caption{
            The overall distribution of the architectures found in each generation.
            The connections from the input layer and the softmax layer are always present and, therefore, they are omitted in the calculation.
        }
        \label{evol} 
    \end{figure*}

    In this section, we will evaluate the proposed architecture regarding its evolution quality and how subpopulations behave throughout the process.
    Figure \ref{evolvable} shows how the mean accuracy of the architectures evolved increases over time.
    The pattern of behaviour is typical of evolutionary algorithms, showing that evolution is happening as expected.
    
    In Figure \ref{evol}, the overall characteristics of the evolved architectures throughout the generations are shown.
    The average number of blocks and the connections between them increase over the generations.
    However, the average number of layers never reaches the same complexity as the initial models.
    The number of layers decreases steeply initially while slowly increasing afterwards.
    Therefore, the overall behaviour is that blocks become smaller and numerous.
    A consequence of this is that the number of connections becomes proportional to the number of connections between blocks and therefore exhibit similar behaviour.
    The average number of layers per block and the average number of connections shows little change, varying only $0.5$ and $0.16$ respectively.
    
    Notice that the average number of layers increases but the average number of layers per block continues to decrease albeit slowly.
    Consequently, blocks tend to degenerate into a few layers, resulting in around three layers per block from the first average number of $3.2$ layers per block.
    Lastly, the average number of connections in a block is kept more or less the same, with the mean varying throughout only from $2.1$ to $2.26$.
    The behaviour described above might suggest that it is hard to create big reusable blocks.
    This seems to be supported by both the decrease of complexity observed as well as the increase in the number of blocks.

\section{Analyzing the Final Architecture: Searching for the Key to Inherent Robustness}
    
    \begin{figure*}[!htb]
    \centering
    \includegraphics[width=0.3\linewidth]{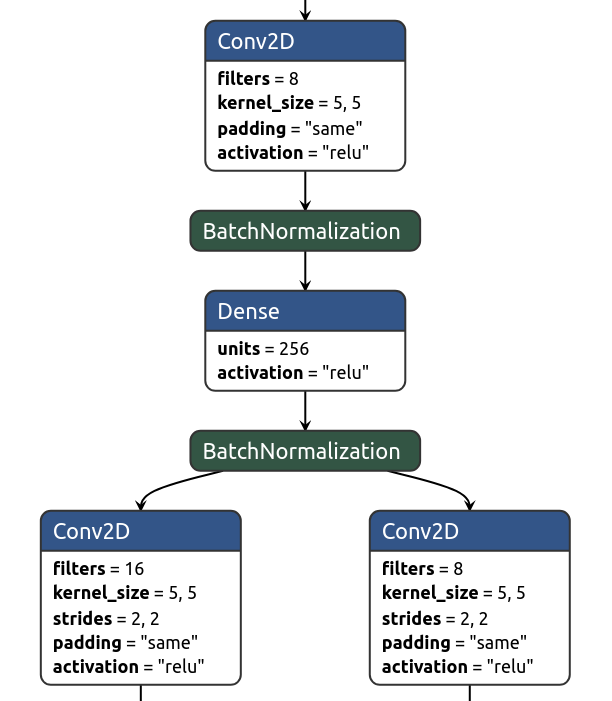} 
    \includegraphics[width=0.48\linewidth]{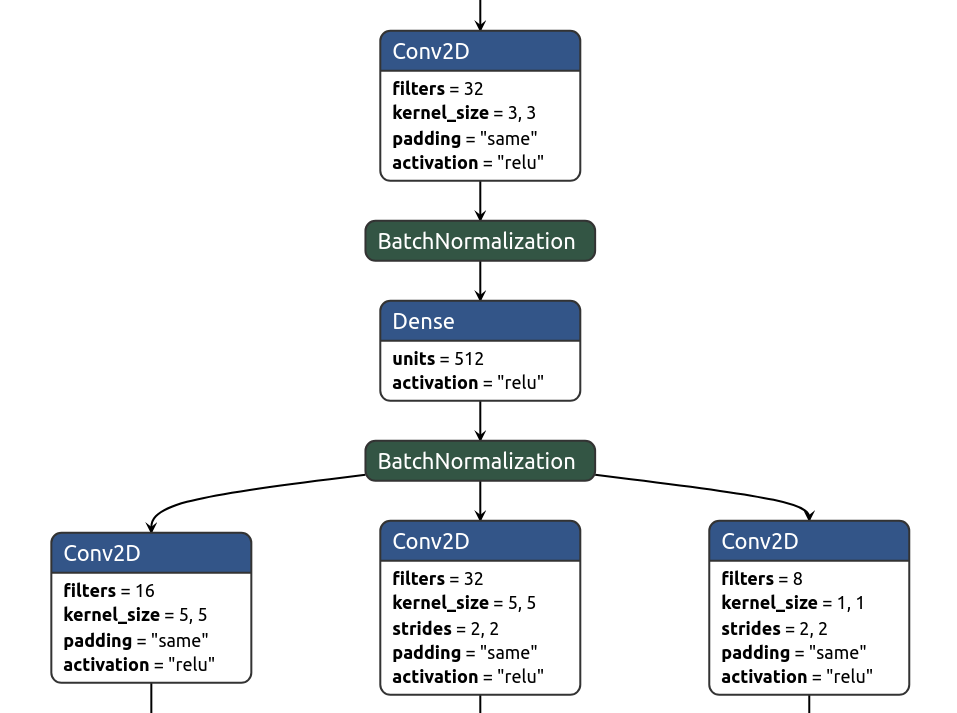} 
    \caption{Two fragments of the evolved architecture which has peculiar traits.}
    \label{peculiar} 
    \end{figure*}

    
    RAS found an architecture that possesses inherent robustness capable of rivalling current defences.
    To investigate the reason behind this robustness, we can take a more in-depth look at the architecture found.
    Figure \ref{peculiar} show(s) some peculiarities from the evolved architecture: multiple bottlenecks, projections into high-dimensional space and paths with different constraints.
    
    \paragraph{Multiple Bottlenecks and Projections into High-Dimensional Space} The first peculiarity is the use of Dense layers in-between Convolutional ones.
    This might seem like a bottleneck similar to the ones used in variational autoencoders.
    However, it is the opposite of a bottleneck (Figure \ref{peculiar}); it is a projection in high-dimensional space.
    The evolved architecture uses mostly a low number of filters while, in some parts of it, high-dimensional projections exist.
    In the whole architecture, four Dense layers in-between Convolutional ones were used, and all of the projects into higher dimensional space.
    This follows directly from Cover's Theorem which states that projecting into high dimensional space makes a training set linearly separable \cite{cover1965geometrical}.

    \paragraph{Paths with Different Constraints} The second peculiarity is the use of multiple paths with the different number of filters and output sizes after high-dimensional projections.
    Notice how the number of filters differs in each of the Convolutional layers in these paths.
    This means there are different constraints over the learning in each of these paths, which should foster different types of features.
    Therefore, this is a multi-bottleneck structure forcing the learning of different sets of features which are now easily constructed from the previous high-dimensional projection.
   
\section{Conclusions}
    
    Automatic search for robust architectures is proposed as a paradigm for developing and researching robust models.
    This paradigm is based on using adversarial attacks together with error rate as evaluation functions in NAS.
    Experiments on using this paradigm with some of the current NAS had poor results.
    This was justified by the small search space used by current methods.
    Here, we propose the RAS method, which has a broader search space, including concatenation, connections between dense to convolutional layer and vice-versa.
    Results with RAS showed that inherently robust architectures do indeed exist.
    In fact, the evolved architecture achieved robust results comparable with state-of-the-art defences while not having any specialised training or defence.
    In other words, the evolved architecture is \textit{inherently robust}.
    Such inherent robustness could increase if adversarial training, or other types of defence, or a combination of them are employed together with it.

    Moreover, investigating the reasons behind such robustness have shown that some peculiar traits are present.
    The evolved architecture has overall a low number of filters and many bottlenecks.
    Multiple projections into high-dimensional space are also present to possibly facilitate the separation of features (Cover's Theorem).
    It also uses multiple paths with different constraints after the high-dimensional projection, which should, consequently, cause a diverse set of features to be learned by the network.
    Thus, in the search space of neural networks, more robust architectures do exist, and more research is required to find and fully document them as well as their features. 
    
\section*{Acknowledgments}

   	This work was supported by JST, ACT-I Grant Number JP-50243 and JSPS KAKENHI Grant Number JP20241216.
    Additionally, we would like to thank Prof. Junichi Murata for the kind support without which it would not be possible to conduct this research.

\bibliographystyle{named}
\bibliography{../adversarial_machine_learning}

\begin{thebibliography}{}

\bibitem[\protect\citeauthoryear{Athalye and
  Sutskever}{2018}]{athalye2017synthesizing}
Anish Athalye and Ilya Sutskever.
\newblock Synthesizing robust adversarial examples.
\newblock In {\em Icml}, 2018.

\bibitem[\protect\citeauthoryear{Athalye \bgroup \em et al.\egroup
  }{2018}]{athalye2018obfuscated}
Anish Athalye, Nicholas Carlini, and David Wagner.
\newblock Obfuscated gradients give a false sense of security: Circumventing
  defenses to adversarial examples.
\newblock In {\em Icml}, 2018.

\bibitem[\protect\citeauthoryear{Brock \bgroup \em et al.\egroup
  }{2017}]{brock2017smash}
Andrew Brock, Theodore Lim, James~M Ritchie, and Nick Weston.
\newblock Smash: one-shot model architecture search through hypernetworks.
\newblock {\em arXiv preprint arXiv:1708.05344}, 2017.

\bibitem[\protect\citeauthoryear{Carlini and Wagner}{2017}]{carlini2017towards}
Nicholas Carlini and David Wagner.
\newblock Towards evaluating the robustness of neural networks.
\newblock In {\em 2017 IEEE Symposium on Security and Privacy (SP)}, pages
  39--57. Ieee, 2017.

\bibitem[\protect\citeauthoryear{Cover}{1965}]{cover1965geometrical}
Thomas~M Cover.
\newblock Geometrical and statistical properties of systems of linear
  inequalities with applications in pattern recognition.
\newblock {\em IEEE transactions on electronic computers}, (3):326--334, 1965.

\bibitem[\protect\citeauthoryear{Goodfellow \bgroup \em et al.\egroup
  }{2014}]{goodfellow2014explaining}
Ian~J Goodfellow, Jonathon Shlens, and Christian Szegedy.
\newblock Explaining and harnessing adversarial examples.
\newblock {\em arXiv preprint arXiv:1412.6572}, 2014.

\bibitem[\protect\citeauthoryear{Hansen \bgroup \em et al.\egroup
  }{2003}]{hansen2003reducing}
Nikolaus Hansen, Sibylle~D M{\"u}ller, and Petros Koumoutsakos.
\newblock Reducing the time complexity of the derandomized evolution strategy
  with covariance matrix adaptation (cma-es).
\newblock {\em Evolutionary computation}, 11(1):1--18, 2003.

\bibitem[\protect\citeauthoryear{He \bgroup \em et al.\egroup
  }{2016}]{he2016deep}
Kaiming He, Xiangyu Zhang, Shaoqing Ren, and Jian Sun.
\newblock Deep residual learning for image recognition.
\newblock In {\em Proceedings of the IEEE conference on computer vision and
  pattern recognition}, pages 770--778, 2016.

\bibitem[\protect\citeauthoryear{He \bgroup \em et al.\egroup
  }{2019}]{he2019automl}
Xin He, Kaiyong Zhao, and Xiaowen Chu.
\newblock Automl: A survey of the state-of-the-art.
\newblock {\em arXiv preprint arXiv:1908.00709}, 2019.

\bibitem[\protect\citeauthoryear{Huang \bgroup \em et al.\egroup
  }{2015}]{huang2015learning}
Ruitong Huang, Bing Xu, Dale Schuurmans, and Csaba Szepesv{\'a}ri.
\newblock Learning with a strong adversary.
\newblock {\em arXiv preprint arXiv:1511.03034}, 2015.

\bibitem[\protect\citeauthoryear{Krizhevsky \bgroup \em et al.\egroup
  }{2009}]{krizhevsky2009learning}
Alex Krizhevsky, Geoffrey Hinton, et~al.
\newblock Learning multiple layers of features from tiny images.
\newblock Technical report, 2009.

\bibitem[\protect\citeauthoryear{Kurakin \bgroup \em et al.\egroup
  }{2016}]{kurakin2016adversarial}
Alexey Kurakin, Ian Goodfellow, and Samy Bengio.
\newblock Adversarial examples in the physical world.
\newblock {\em arXiv preprint arXiv:1607.02533}, 2016.

\bibitem[\protect\citeauthoryear{Lee \bgroup \em et al.\egroup
  }{2018}]{lee2018hardness}
Hyeon-Chang Lee, Dong-Pil Yu, and Yong-Hyuk Kim.
\newblock On the hardness of parameter optimization of convolution neural
  networks using genetic algorithm and machine learning.
\newblock In {\em Proceedings of the Genetic and Evolutionary Computation
  Conference Companion}, pages 51--52, 2018.

\bibitem[\protect\citeauthoryear{Lima and Pozo}{2019}]{lima2019evolving}
Ricardo~HR Lima and Aurora~TR Pozo.
\newblock Evolving convolutional neural networks through grammatical evolution.
\newblock In {\em Proceedings of the Genetic and Evolutionary Computation
  Conference Companion}, pages 179--180, 2019.

\bibitem[\protect\citeauthoryear{Madry \bgroup \em et al.\egroup
  }{2018}]{madry2017towards}
Aleksander Madry, Aleksandar Makelov, Ludwig Schmidt, Dimitris Tsipras, and
  Adrian Vladu.
\newblock Towards deep learning models resistant to adversarial attacks.
\newblock In {\em Iclr}, 2018.

\bibitem[\protect\citeauthoryear{Miikkulainen \bgroup \em et al.\egroup
  }{2019}]{miikkulainen2019evolving}
Risto Miikkulainen, Jason Liang, Elliot Meyerson, Aditya Rawal, Daniel Fink,
  Olivier Francon, Bala Raju, Hormoz Shahrzad, Arshak Navruzyan, Nigel Duffy,
  et~al.
\newblock Evolving deep neural networks.
\newblock In {\em Artificial Intelligence in the Age of Neural Networks and
  Brain Computing}, pages 293--312. Elsevier, 2019.

\bibitem[\protect\citeauthoryear{Moosavi-Dezfooli \bgroup \em et al.\egroup
  }{2017}]{moosavi2017universal}
Seyed-Mohsen Moosavi-Dezfooli, Alhussein Fawzi, Omar Fawzi, and Pascal
  Frossard.
\newblock Universal adversarial perturbations.
\newblock In {\em 2017 IEEE Conference on Computer Vision and Pattern
  Recognition (CVPR)}, pages 86--94. Ieee, 2017.

\bibitem[\protect\citeauthoryear{Negrinho and
  Gordon}{2017}]{negrinho2017deeparchitect}
Renato Negrinho and Geoff Gordon.
\newblock Deeparchitect: Automatically designing and training deep
  architectures.
\newblock {\em arXiv preprint arXiv:1704.08792}, 2017.

\bibitem[\protect\citeauthoryear{Papernot \bgroup \em et al.\egroup
  }{2016}]{papernot2016distillation}
Nicolas Papernot, Patrick McDaniel, Xi~Wu, Somesh Jha, and Ananthram Swami.
\newblock Distillation as a defense to adversarial perturbations against deep
  neural networks.
\newblock In {\em 2016 IEEE Symposium on Security and Privacy (SP)}, pages
  582--597. Ieee, 2016.

\bibitem[\protect\citeauthoryear{Pham \bgroup \em et al.\egroup
  }{2018}]{pham2018efficient}
Hieu Pham, Melody Guan, Barret Zoph, Quoc Le, and Jeff Dean.
\newblock Efficient neural architecture search via parameter sharing.
\newblock In {\em International Conference on Machine Learning}, pages
  4092--4101, 2018.

\bibitem[\protect\citeauthoryear{Real \bgroup \em et al.\egroup
  }{2017}]{real2017large}
Esteban Real, Sherry Moore, Andrew Selle, Saurabh Saxena, Yutaka~Leon Suematsu,
  Jie Tan, Quoc~V Le, and Alexey Kurakin.
\newblock Large-scale evolution of image classifiers.
\newblock In {\em Proceedings of the 34th International Conference on Machine
  Learning-Volume 70}, pages 2902--2911. JMLR. org, 2017.

\bibitem[\protect\citeauthoryear{Sabour \bgroup \em et al.\egroup
  }{2017}]{sabour2017dynamic}
Sara Sabour, Nicholas Frosst, and Geoffrey~E Hinton.
\newblock Dynamic routing between capsules.
\newblock In {\em Advances in neural information processing systems}, pages
  3856--3866, 2017.

\bibitem[\protect\citeauthoryear{Sharif \bgroup \em et al.\egroup
  }{2016}]{sharif2016accessorize}
Mahmood Sharif, Sruti Bhagavatula, Lujo Bauer, and Michael~K Reiter.
\newblock Accessorize to a crime: Real and stealthy attacks on state-of-the-art
  face recognition.
\newblock In {\em Proceedings of the 2016 ACM SIGSAC Conference on Computer and
  Communications Security}, pages 1528--1540. Acm, 2016.

\bibitem[\protect\citeauthoryear{Storn and Price}{1997}]{storn1997differential}
R.~Storn and K.~Price.
\newblock Differential evolution--a simple and efficient heuristic for global
  optimization over continuous spaces.
\newblock {\em Journal of global optimization}, 11(4):341--359, 1997.

\bibitem[\protect\citeauthoryear{Su \bgroup \em et al.\egroup
  }{2019}]{su2019one}
Jiawei Su, Danilo~Vasconcellos Vargas, and Kouichi Sakurai.
\newblock One pixel attack for fooling deep neural networks.
\newblock {\em IEEE Transactions on Evolutionary Computation}, 23(5):828--841,
  2019.

\bibitem[\protect\citeauthoryear{Szegedy}{2014}]{szegedy2014intriguing}
Christian et~al. Szegedy.
\newblock Intriguing properties of neural networks.
\newblock In {\em In ICLR}. Citeseer, 2014.

\bibitem[\protect\citeauthoryear{Uesato \bgroup \em et al.\egroup
  }{2018}]{uesato2018adversarial}
Jonathan Uesato, Brendan O'Donoghue, Pushmeet Kohli, and Aaron Oord.
\newblock Adversarial risk and the dangers of evaluating against weak attacks.
\newblock In {\em International Conference on Machine Learning}, pages
  5032--5041, 2018.

\bibitem[\protect\citeauthoryear{Vargas and
  Kotyan}{2019}]{vargas2019robustness}
Danilo~Vasconcellos Vargas and Shashank Kotyan.
\newblock Robustness assessment for adversarial machine learning: Problems,
  solutions and a survey of current neural networks and defenses.
\newblock {\em arXiv preprint arXiv:1906.06026}, 2019.

\bibitem[\protect\citeauthoryear{Vargas and Murata}{2017}]{vargas2017spectrum}
Danilo~Vasconcellos Vargas and Junichi Murata.
\newblock Spectrum-diverse neuroevolution with unified neural models.
\newblock {\em IEEE transactions on neural networks and learning systems},
  28(8):1759--1773, 2017.

\bibitem[\protect\citeauthoryear{Vargas and Su}{2019}]{vargas2019understanding}
Danilo~Vasconcellos Vargas and Jiawei Su.
\newblock Understanding the one-pixel attack: Propagation maps and locality
  analysis.
\newblock {\em arXiv preprint arXiv:1902.02947}, 2019.

\bibitem[\protect\citeauthoryear{Xu \bgroup \em et al.\egroup
  }{2017}]{xu2017feature}
Weilin Xu, David Evans, and Yanjun Qi.
\newblock Feature squeezing: Detecting adversarial examples in deep neural
  networks.
\newblock {\em arXiv preprint arXiv:1704.01155}, 2017.

\bibitem[\protect\citeauthoryear{Yu and Kim}{2018}]{yu2018worth}
Dong-Pil Yu and Yong-Hyuk Kim.
\newblock Is it worth to approximate fitness by machine learning? investigation
  on the extensibility according to problem size.
\newblock In {\em Proceedings of the Genetic and Evolutionary Computation
  Conference Companion}, pages 77--78, 2018.

\bibitem[\protect\citeauthoryear{Zoph \bgroup \em et al.\egroup
  }{2018}]{zoph2018learning}
Barret Zoph, Vijay Vasudevan, Jonathon Shlens, and Quoc~V Le.
\newblock Learning transferable architectures for scalable image recognition.
\newblock In {\em Proceedings of the IEEE conference on computer vision and
  pattern recognition}, pages 8697--8710, 2018.

\end{thebibliography}

\end{document}